%% file: _main.tex
\input{_constants}
\arxiv
\documentclass[10pt,twocolumn,letterpaper]{article}
\usepackage{tikz}
\usetikzlibrary{matrix}
\usetikzlibrary{decorations.pathreplacing}
\input{cvpr_header}
\begin{document}
%% TITLE
\title{\paperTitle}
\author{\authorBlock}
\maketitle

\input{01_main_text}

{\small
\bibliographystyle{ieee_fullname}
\bibliography{11_references}
}

% \ifarxiv \clearpage \input{12_appendix} \fi
\clearpage \input{12_appendix}

\end{document}

%% file: _constants.tex
\def\cvprPaperID{0000}
\def\confName{ICCV}
\def\confYear{2023}

\def\paperTitle{TeleViT: Teleconnection-driven Transformers Improve Subseasonal to Seasonal Wildfire Forecasting}

\def\authorBlock{

    {Ioannis Prapas}\textsuperscript{1, 3} \qquad
    {Nikolaos Ioannis Bountos}\textsuperscript{1,2} \qquad
    Spyros Kondylatos\textsuperscript{1, 3} \qquad \\ 
    Dimitrios Michail\textsuperscript{2} \qquad 
    Gustau Camps-Valls\textsuperscript{3} \qquad
    Ioannis Papoutsis\textsuperscript{1} \\
    \textsuperscript{1}\textit{Orion Lab, Institute for Astronomy, Astrophysics, Space Applications, } \\ 
    \textit{and Remote Sensing (IAASARS), National Observatory of Athens} \\
    \textsuperscript{2}\textit{Department of Informatics and Telematics, Harokopio University of Athens} \\
    \textsuperscript{3}\textit{Image Processing Laboratory (IPL), Universitat de Val\`encia}
}

\newif\ifreview 
\newif\ifarxiv \newcommand{\arxiv}{\arxivtrue}
\newif\ifcamera 
\newif\ifrebuttal 

%% file: cvpr_header.tex
\ifreview \usepackage[review]{cvpr} \fi
\ifarxiv \usepackage[pagenumbers]{cvpr} \fi
\ifrebuttal \usepackage[rebuttal]{cvpr} \fi
\ifcamera \usepackage{cvpr} \fi

\usepackage{graphicx}
\usepackage{amsmath}
\usepackage{amssymb}
\usepackage{booktabs}

\input{_macros}  % Add packages to _macros.tex

\usepackage{xr-hyper}

\makeatletter
\newcommand*{\addFileDependency}[1]{
  \typeout{(#1)}
  \@addtofilelist{#1}
  \IfFileExists{#1}{}{\typeout{No file #1.}}
}

\makeatother

\usepackage[pagebackref,breaklinks,colorlinks]{hyperref}
\usepackage[capitalize]{cleveref}
\crefname{section}{Sec.}{Secs.}
\crefname{table}{Table}{Tables}
\crefname{figure}{Fig.}{Figs.}

%% file: _macros.tex
\usepackage{times}
\usepackage{microtype}
\usepackage{epsfig}
\usepackage{caption}
\usepackage{float}
\usepackage{placeins}
\usepackage{color, colortbl}
\usepackage{stfloats}
\usepackage{enumitem}
\usepackage{tabularx}
\usepackage{xstring}
\usepackage{multirow}
\usepackage{xspace}
\usepackage{url}
\usepackage{subcaption}
\usepackage{xcolor}
\usepackage[hang,flushmargin]{footmisc}

\ifcamera \usepackage[accsupp]{axessibility} \fi

% ...

\ifarxiv  \fi

\newcommand{\R}[1]{{%
    \textbf{%
        \ifstrequal{#1}{1}{\textcolor{red}{R#1}}{%
        \ifstrequal{#1}{2}{\textcolor{blue}{R#1}}{%
        \ifstrequal{#1}{3}{\textcolor{magenta}{R#1}}{%
        \ifstrequal{#1}{4}{\textcolor{teal}{R#1}}{%
                           \textcolor{cyan}{R#1}%
        }}}}%
    }%
}}

%% file: 01_main_text.tex
\begin{abstract}
% Abstract goes here.

Wildfires are increasingly exacerbated as a result of climate change, necessitating advanced proactive measures for effective mitigation. It is important to forecast wildfires weeks and months in advance to plan forest fuel management, resource procurement and allocation. To achieve such accurate long-term forecasts at a global scale, it is crucial to employ models that account for the Earth system's inherent spatio-temporal interactions, such as memory effects and teleconnections. We propose a teleconnection-driven vision transformer (TeleViT), capable of treating the Earth as one interconnected system, integrating fine-grained local-scale inputs with global-scale inputs, such as climate indices and coarse-grained global variables. Through comprehensive experimentation, we demonstrate the superiority of TeleViT in accurately predicting global burned area patterns for various forecasting windows, up to four months in advance. The gain is especially pronounced in larger forecasting windows, demonstrating the improved ability of deep learning models that exploit teleconnections to capture Earth system dynamics. Code available at \href{https://github.com/Orion-Ai-Lab/TeleViT}{github.com/Orion-Ai-Lab/TeleViT}.

\end{abstract}

\section{Introduction}
\label{sec:introduction}

% MOTIVATION
% \paragraph{Motivation} 
Global warming increases the frequency and intensity of fire weather, amplifying the likelihood of the conditions that lead to extreme wildfire events \cite{jones_global_2022}. %Extreme wildfire events occur in conditions that are becoming more probable with climate change.
In that context, it is important to improve our understanding of wildfires, anticipating wildfire patterns weeks and months in advance. Operationally, subseasonal to seasonal wildfire forecasts are typically treated as anticipated anomalies in temperature and precipitation, derived from process-based climate models \cite{effis_seasonal_forecast}, while disregarding crucial fire-related factors like soil moisture and vegetation dynamics. Deep Learning (DL) methods, able to learn from data, offer a promising avenue to model Earth system processes such as wildfires \cite{reichstein_deep_2019}, arising from the dynamic interactions of all the different fire drivers, namely climate, vegetation and human activity \cite{hantson_status_2016}.
% Given the disastrous effects of wildfires, it is of utmost importance to provide wildfire danger forecasts weeks or even months in advance.
% Improving our understanding of the wildfire phenomenon is crucial for this goal.
% Deep Learning approaches, with their ability to learn from data,

% PREVIOUS WORK
% Deep learning methods have seen remarkable success as data-driven learners, offering a promising avenue to modeling dynamic Earth system processes \cite{reichstein_deep_2019} and in particular, identifying the sufficient conditions that could lead to the ignition of wildfires. %, arising from the dynamic interactions of the fire drivers, namely climate, vegetation and human activity \cite{hantson_status_2016}.
\paragraph{Previous work} In fact, several studies have successfully applied DL to wildfire forecasting tasks \cite{jain_review_2020,ghali_deep_2023}. When it comes to global predictions on subseasonal to seasonal scales, existing work relies on traditional Machine Learning (ML) approaches \cite{prapas_deep_2022} that do not effectively capture Earth system dynamics that are important for long-term forecasting, even if sometimes they use teleconnection indices as input features \cite{li_attentionfire_v10_2023, yu_quantifying_2020}. In this work, we argue that for predictions on long temporal scales, it is crucial to train models that consider the Earth as one interconnected system, accounting for spatio-temporal interactions such as memory effects and teleconnections. There is substantial evidence that teleconnections modulate global wildfires \cite{justino_arctic_2022, kim_extensive_2020, forkel_extreme_2012, cardil_climate_2023}.
For example, extreme wildfires in Siberia have been linked to preceding arctic oscillation \cite{kim_extensive_2020} and previous-year soil moisture anomalies \cite{forkel_extreme_2012}. Despite the evidence, there is very limited ML work to predict wildfires that combines local information from the fire drivers with Earth system variables. Chen \etal \cite{chen_forecasting_2020} use simple autoregressive statistical methods to combine volume pressure deficit values with oceanic indicators. Yu \etal \cite{yu_quantifying_2020} use a statistical pre-processing to identify the most prominent oceanic indicators modulating burned area in Africa and then use the findings to select input features for tree-based machine learning models. AttentionFire \cite{li_attentionfire_v10_2023} uses an attention-enhanced recurrent neural network architecture that considers temporal context and shows slight improvements at longer horizons when adding information from oceanic indices. More sophisticated models that handle the Earth as a system are arising naturally for weather and climate prediction. Earthformer \cite{gao_Earthformer_2022} combines spatio-temporal attention with learnable global vectors that are meant to summarize the dynamics of a system. GraphCast \cite{lam_graphcast_2022} uses graph neural networks that capture long-range interactions by defining graphs at several resolutions. Climax \cite{nguyen_climax_2023} is proposed as a transformer-based foundational model for weather and climate at coarse resolutions, working with global input or on a very large spatial domain.

\paragraph{Application context} 
Our study has direct implications for humanitarian assistance and disaster response (HADR) operations, particularly in addressing the challenges of intensified wildfires due to climate change. Forecasting wildfires weeks or months in advance can inform the deployment of HADR-related anticipatory actions.
This includes measures for vegetation and forest management (\eg controlled burns and targeted firebreaks), coordination with local authorities to form evacuation plans, procurement of resources such as leasing firefighting equipment, and timely mobilisation of international and cross-border cooperation and aid.

Our transformer-based models combine local and non-local inputs like teleconnection indices and coarsened global variables, and thus offer a novel approach to model Earth system processes and improve our capabilities to forecast wildfires at subseasonal to seasonal scales. 
Outside of wildfire forecasting, these methods hold promise for diverse HADR applications, enabling foresight across extended timeframes, which can empower proactive planning and optimized resource allocation before disastrous events.

% Contributions
\paragraph{Contributions} In this work, we propose to explicitly model short (local) and long (global) range interactions of the Earth system for long-term global wildfire pattern forecasting. 
For this, we develop a Teleconnection-driven Vision Transformer (TeleViT), that expands ViT \cite{dosovitskiy_image_2021} with an asymmetric tokenisation procedure in order to seamlessly combine local and global scale inputs, \ie climate indices and coarsened global variables. 
We thoroughly examine the performance of the proposed model, demonstrating superior performance in various forecasting windows compared to models that do not leverage teleconnections. 

\section{Teleconnection Vision Transformer (TeleViT)}
\label{sec:method}

To produce reliable long-range forecasts (weeks or even months in advance), it is imperative to treat the Earth as a connected system. The Earth system is characterised by continuous interactions of processes that spread over large spatio-temporal windows, manifesting as memory effects and teleconnections.
% When making predictions weeks or months in advance, it is important to consider the Earth as a system. This means taking into account its large-scale spatio-temporal interactions, that manifest as memory effects and teleconnections explained below. 
\textit{Memory effects} are the realisation of the persistence or influence of past events on current and future states. They represent the temporal aspect of system behaviour, where the history of a system influences its present or future behaviour. In the context of wildfire prediction, memory effects capture how past events such as fuel accumulation, drought conditions, and weather patterns can impact the likelihood, behaviour, and extent of wildfires.
\textit{Teleconnections} constitute long-distance interactions between different regions in the Earth system. They describe how changes in one region can influence atmospheric or oceanic conditions in another, often through large-scale atmospheric circulation patterns or oceanic phenomena \cite{wallace_teleconnections_1981}. Teleconnections are often described in terms of teleconnection indices, \ie Oceanic and Climatic Indices (OCIs), calculated as large-scale anomalies of specific parameters, such as temperature, pressure or sea surface temperature. By teleconnections, however, we refer to the long-range spatio-temporal interactions and not to the indices per se, which we consider a mere proxy to the state %and dynamics 
of the Earth system.
% Teleconnections can lead to spatially coherent patterns of climate variability and can have implications for weather patterns, climate anomalies, and the occurrence of extreme wildfire events. 

% Use figure* for multi-column figure
\begin{figure}[htbp]
    \centering
    \includegraphics[width=\linewidth]{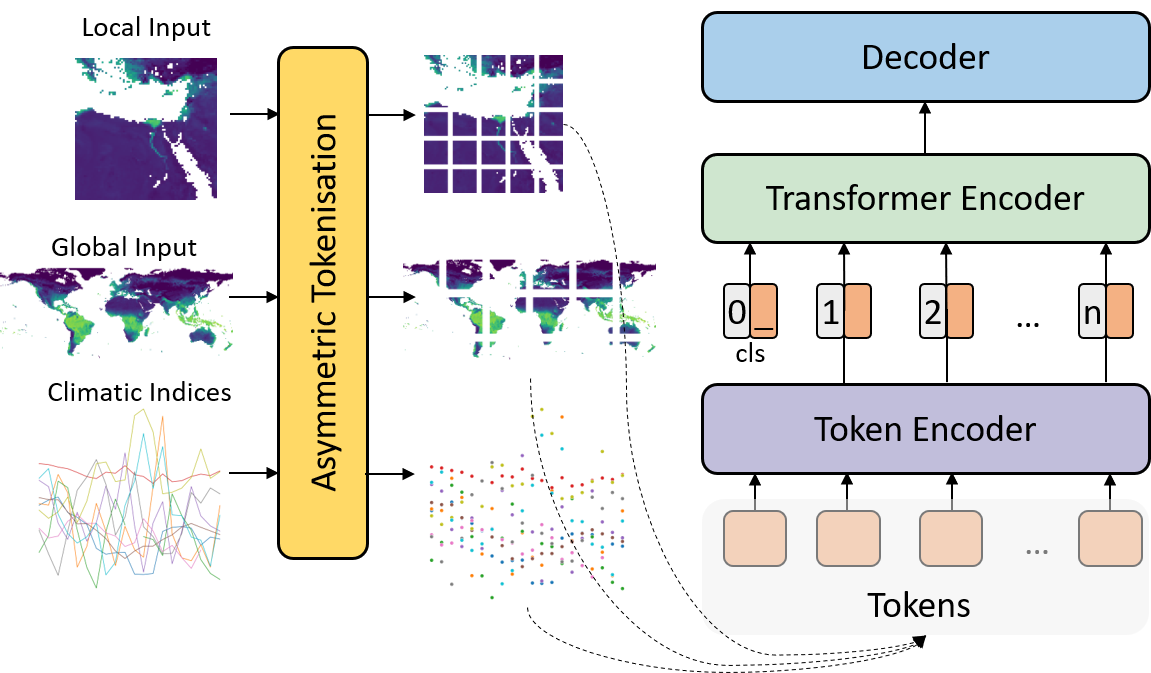}
    \caption{Full pipeline of the TeleViT architecture. The different multi-scale inputs \ie local, global and teleconnection indices, are tokenized at different resolutions and fed to a Transformer encoder along with a prepended classification token. The linear decoder is based on the output of the classification token.}
    \label{fig:esdl}
\end{figure}

% To insert a figure: \input{figs/template}
% Or table: \input{tables/template}

We propose to capture such distant interactions that are omnipresent in the Earth system, using data that can inform on the state and dynamics of the system. Particularly, we propose to enhance fine-resolution local data, \ie information for a small area on Earth, that are commonly used in isolation, with information from i) OCIs and ii) coarse-resolution global data.
Effectively combining these inputs is particularly challenging due to high dimensionality and the discrepancy between the spatial resolution of the different data sources.

In that direction, we propose to utilise the versatility of the Transformer architecture\cite{vaswani_attention_2017}. We build upon the Vision Transformer (ViT) \cite{dosovitskiy_image_2021}, which adapts the architecture to computer vision problems by splitting an image into non-overlapping chips (tokens). The resulting token sequence serves as input to a standard Transformer. Extending ViT, we introduce the \textit{Teleconnection-driven Vision Transformer (TeleViT)}. TeleViT relies on an asymmetric tokenization method which increments the sequence of input tokens with tokens stemming from different data sources, with potentially varying spatial and temporal resolutions. To produce the token sequence, each input source is tokenised independently, taking into account its inherent characteristics along with the scale it operates on the Earth system. Local information and time series of climatic indices should be represented in high detail with smaller token sizes, while global information, operating at a greater scale, can be less detailed benefitting from larger token sizes. 
% We see the different inputs tokenised independently and then passed to the Transformer architecture. We can choose to include or not any of the input sources. 
% 

TeleViT can be adapted to consume any type of input, given a tokenisation strategy. For our setting, we assume local input $x_l \in R^{C_l\times H_l\times W_l}$, global input $x_g\in R^{C_{g}\times H_{g}\times W_{g}}$ and indices input $x_{i}\in R^{C_i \times T}$, 
with C the number of features, H the height, W the width and T the time-series length.
According to our asymmetric tokenisation, each of the inputs  is split into tokens individually, resulting in $N_l$, $N_g$ and $N_i$ tokens with dimensions $P_l$, $P_g$ and $P_i$ that depend on the selected tokenisation strategy. %The Transformer uses constant latent vector size D through all of its layers, so we map the tokens to D dimensions with a different trainable linear projection function for each input $f_k: R^{N_k \times P_k} \rightarrow R^{N_k \times D}, k \in {l, g, i}$.
Tokens are mapped to the embedding dimension $D$ of the Transformer with a different trainable linear projection function for each input $f_k: R^{N_k \times P_k} \rightarrow R^{N_k \times D}, k \in {l, g, i}$. 
% , matching the constant latent vector of size D, used in all of the Transformer's layers.
Similar to ViT, we add learnable positional encoding to the sequence of $N_l + N_g + N_i$ number of tokens, which is fed to a standard Transformer with depth $K$ and $A$ attention heads for each layer. Pixel-level prediction is performed by a simple decoder attached to the prepended classification token (cls). We choose a trainable linear layer that maps to the input resolution. The architecture is demonstrated in Figure \ref{fig:esdl}.

% 
% with $C_V$, $H_V$ and $W_V$ the number of input variables, the height and the width of each dataset $V$, as well as oceanic indices $I\in R^{M\times T}$, where $M$ and $T$ the number of independent oceanic indices and their temporal dimension respectively. Each dataset $V$ is assigned a unique tokenising function $f_V(p_V)$, parameterized by a dataset-specific patch size $p_V$ that creates a sequence of $N_V$ non-overlapping patches, linearly projected to a $D$-dimensional space. The outputs of the independent tokenisers are concatenated to form a sequence $[S_L; S_G; S_{I}]$, where $S_{V}\in R^{N\times D}$  $\forall V \in \{L, G, I\}$. The resulting sequence is fed to a standard Transformer with depth $K$ and $A$ attention heads for each layer.

 % TeleViT fosters the combination of information between local high-resolution patches from the region of interest, with global-scale views and OCIs. 
 
 As the Transformer architecture remains intact, most of the inductive biases come from the tokenisation procedure. By keeping the inductive bias to a minimum we remove any restrictions to known concepts, enabling 
 unprecedented information combination of different input datasets, representing local and global inputs. 
 For example, attention allows i) inter-dataset interactions, \ie information flows from distant large-scale regions of coarse resolution to high-resolution localised windows, and ii) intra-dataset interactions, where several Earth system processes and interactions are modelled at a global scale while identifying location-specific characteristics for each region independently (depicted in Figure \ref{fig:attn_multires}, Appendix \ref{app:attn_multires}). The simplicity of the method makes it easily extensible to other data sources, as well as to the inclusion of a temporal dimension that we do not address in this study.

\section{Experiments}
\label{sec:experiments}
We conduct our experiments on the SeasFire cube \cite{alonso_seasfire_2022}, a spatio-temporal dataset for subseasonal to seasonal wildfire forecasting. It contains 21 years of data (2001-2021) at a global scale, in an 8-days temporal and $0.25^{\circ}$  spatial resolution. The cube includes a diverse range of seasonal fire drivers, combining atmospheric, vegetation, and anthropogenic variables along with climate indices, in addition to target variables related to wildfires such as burned areas, fire radiative power, and wildfire-related emissions. 

Building on the SeasFire cube, Prapas \etal\cite{prapas_deep_2022} defined burned area pattern forecasting as a segmentation task. As input, they use local patches that contain different channels of the fire driver variables and train a U-Net++\cite{zhou_unet_2020} to predict the presence of burned areas at a future time step. Their U-Net++ demonstrated a predictive skill greater than the burned area mean seasonal cycle for a lead forecasting time of up to $2$ months. We follow a similar setup, where given a snapshot of the fire driver variables at timestep $t$ we want to predict the presence of burned areas at a future timestep $t+h$. For the local input, we extract patches of size $80\times80$, and as such, the world represented at $0.25^{\circ}$ %\textdegree
($1440\times720$ cells), is split into $18\times9 = 162$ local input patches. To extract the global input, we coarsen the cube to $1^{\circ}$ (see Appendix \ref{app:coarsening}), reducing its size by a factor of 16, making the global input size 360$\times$180. Along with 10 fire driver variables extracted from the cube, the same for both local and global inputs, we calculate a \textit{global positional encoding}, \ie sine and cosine of the longitude and latitude. For each sample, 10 OCIs are extracted for the 10 months preceding $t$. The target is the presence of burned area at time-step $t+h$ for the region of the local input, where h is the lead time forecasting horizon. As such, a sample is comprised of four vectors; i) a local input $x_l$ of size $(14, 80, 80)$, ii) a global input $x_g$ of size $(14, 360, 180)$, iii) an OCI input $x_i$ of size $(10, 10)$ and iv) a target vector of size $(1, 80, 80)$. The variables used are shown in Appendix \ref{app:datacube}. The model's performance is evaluated using the Area Under the Precision-Recall Curve (AUPRC). The train, validation, and test split is time-based, using years 2002 - 2017 for training, 2018 for validation and 2019 for testing. Only samples that contain burned areas are considered. % We evaluate our architecture in three different settings: a) using only local information, b) combining local information and teleconnections indices and c) including local information, teleconnection indices and global scale information.%

We assess the following models: i) a U-Net++, which uses only local input $x_l$ \cite{prapas_deep_2022}, ii)  a simple ViT, which uses only local input $x_l$, iii) $\mathrm{TeleViT}_{i}$, which uses OCIs $x_i$ along with local input $x_l$,
iv) $\mathrm{TeleViT}_{g}$, which uses only global input $x_g$ along with local input $x_l$, and
v) $\mathrm{TeleViT}_{i,g}$, which uses both OCIs $x_i$ and global input $x_g$ along with the local input $x_l$.
The hyper-parameters have been tuned for simple ViT and applied to TeleViT models that use the same core architecture. We provide detailed information on architectural and training choices in Appendix \ref{app:hparams}.

% We compare our proposed method with the standard U-Net++ and ViT architectures. TeleViT is built upon ViT's architecture and is examined with all possible input combinations. For clarity, we indicate the input sources used in each TeleViT model as $\mathrm{TeleViT}_{i}$, $\mathrm{TeleViT}_{g}$ and $\mathrm{TeleViT}_{i,g}$ with $i$ and ${g}$ representing the indices and global inputs respectively, as defined in section \ref{sec:method}. U-Net++ and ViT are trained using solely the local inputs. 

The performance of the five models is examined in several forecasting horizons $h \in \{1, 2, 4, 8, 16\}$. %, \ie models are trained to estimate the burned area pattern of the target 8-day period using input data from 1, 2, 4, 8 and 16 number of 8-days ahead of time.
A model that predicts at the maximum forecasting horizon, \ie 16$\times$8-days in advance, learns to predict the burned area pattern of a particular 8-day period approximately four months in advance. A different model is trained for each $h$, for a total of $25$ experiments.

% All models are trained to forecast the burned area at 5 forecasting lead times, \ie 1, 2, 4, 8 8-days in advance.

\section{Results and Discussion}
% The further one attempts to see into the future, the more uncertain the predictions become. 
% Interestingly enough, our top performing model achieves similar performance at the 16 8-days window as our baseline at the simple 1 8-day, showcasing the prowess of our method.

    % \item $\mathrm{TeleViT}_{ci}$, which uses both local input and climatic indices input.
    % \item $\mathrm{TeleViT}_{g}$, which uses coarsened global views along with local input.
    % \item $\mathrm{TeleViT}_{ci,g}$,

% Use figure* for multi-column figure
%between the two.

\begin{figure}[t]
    \centering
    \includegraphics[width=\linewidth]{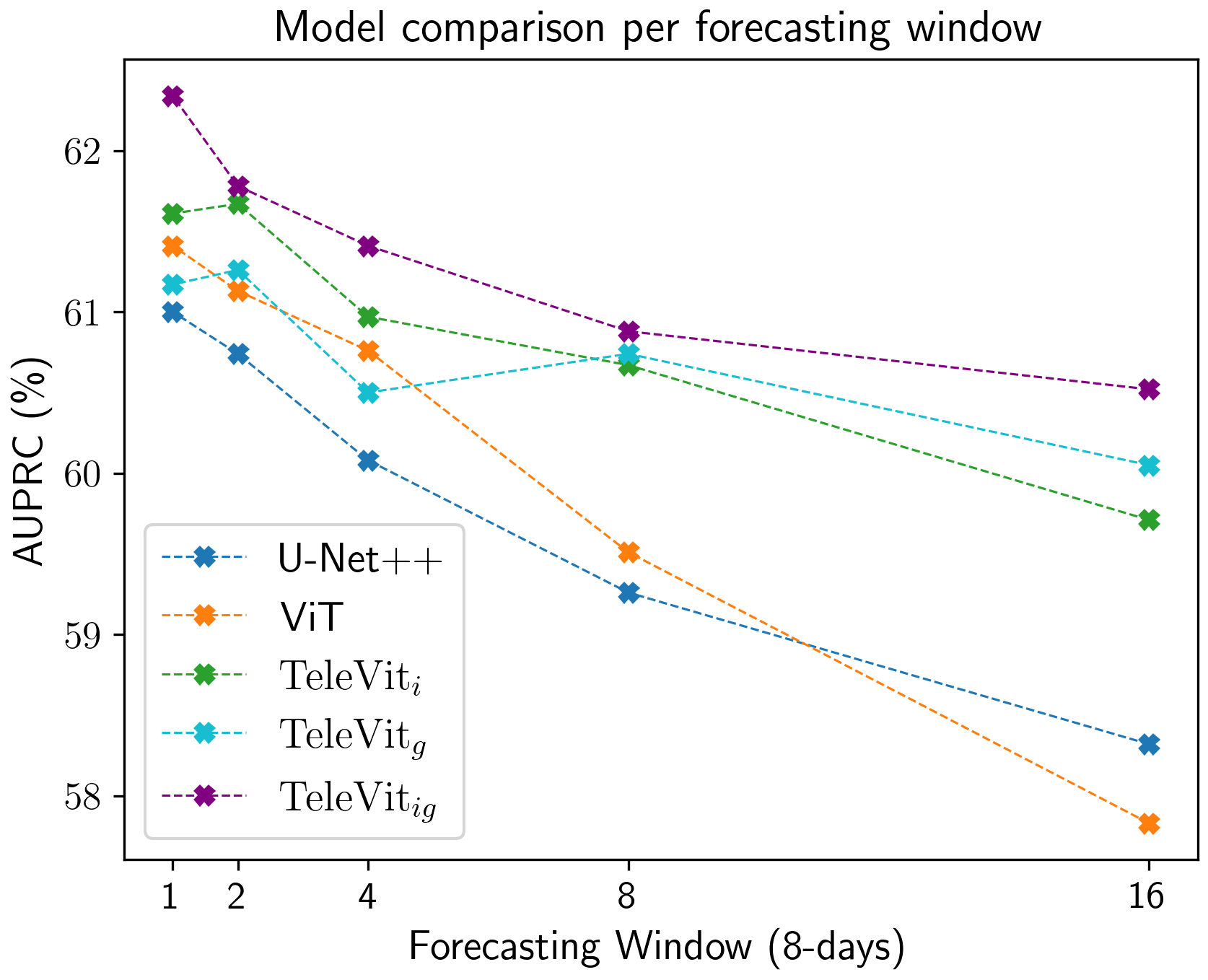}
    \caption{AUPRC performance of the different models for forecasting windows of 1, 2, 4, 8 and 16$\times$8-days in advance.}
    \label{fig:experiments}
\end{figure}

\begin{figure}[htbp]
    \centering
    \includegraphics[width=\linewidth]{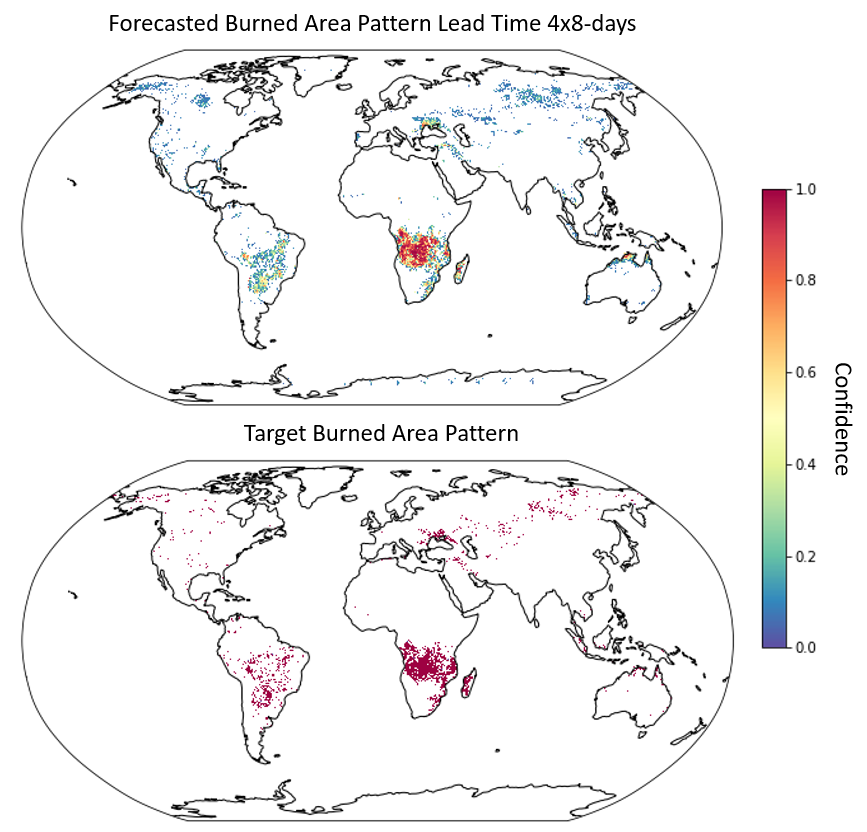}
    \caption{A sample prediction versus the target for the best model, predicting at 4$\times$8-days lead forecasting time. Sea and values lower than 0.05 are masked out. Confidence is determined as the softmax score of the positive prediction.}
    \label{fig:danger_maps}
\end{figure}

Figure \ref{fig:experiments} summarises the results of the experiments.  
In general, all the models exhibit a declining trend in performance as the forecasting window increases. The decline, however, is much less steep for teleconnection-driven models, than for the baselines (ViT and U-Net++). ViT proves to be a stronger baseline, outperforming U-Net++ with the exception of the 16$\times$8-days, where U-Net++ overtakes it by a small margin. $\mathrm{TeleViT}_{g}$ achieves a comparable performance to ViT for short forecasting windows of up to 4$\times$8-days, while it shows greater robustness to the increase of the forecasting window. $\mathrm{TeleViT}_{i}$ and $\mathrm{TeleViT}_{i,g}$ consistently surpass the baselines, with larger performance gaps as the lead time increases. Notably, $\mathrm{TeleViT}_{i,g}$  has the dominant performance, which indicates the benefits of a synergistic effect between teleconnection indices and global-scale representations. Interestingly, it achieves high gains even in short forecasting windows, which suggests that contextual information brought by this combination is helpful beyond teleconnections that operate on larger temporal scales. Figure \ref{fig:danger_maps} shows how the predictions compare to the target for a specific sample date, demonstrating high agreement.

% Interestingly, $\mathrm{TeleViT}_{i,g}$ gives a pronounced boost in performance also in short forecasting windows, which suggests a synergistic effect between teleconnection indicces and global-scale representations.
% showcasing its ability to utilise global context, especially in greater forecasting horizons where the impact of teleconnections is expected to be more significant. 
% Interestingly, TeleViT exhibits improved performance even in shorter forecasting windows, which suggests that contextual information is helpful beyond teleconnections that operate on larger temporal scales. When we compare $\mathrm{TeleViT}_{i}$ and $\mathrm{TeleViT}_{g}$, we notice that for shorter forecasting horizons the former performs better. This pattern is inverted as the forecasting window grows larger.

% A significant result from our experiments is the ability of $\mathrm{TeleViT}_{i,g}$ to build on top of $\mathrm{TeleViT}_{i}$ and $\mathrm{TeleViT}_{g}$, strictly outperforming both of them. This proves TeleViT's ability to fuse information from multiple sources effectively, opening interesting research questions towards the applicability of our method to other Earth Science-related tasks. from multiple sources effectively, opening interesting research questions

In general, the results demonstrate TeleViT's ability to successfully fuse local and non-local Earth system information and pave the road for exciting future research.
% , bringing multiple questions to the foreground. 
Both global-scale inputs and OCIs can bring performance gains, but it is not clear if this is due to the same reasons. Further work is needed to investigate the relative contribution of each input. This urges us to comprehend the models, elucidating both familiar and undiscovered interactions between teleconnections and their influence on burned area patterns. A thorough examination of the attention maps may offer valuable insights in this regard, potentially shedding light on the underlying mechanisms.
% Moreover, given the success of the proposed tokenisation method, t
There is also much potential for further investigation of tokenisation schemes. The performance improvement induced by the introduction of coarsened global views may suggest that OCIs are a simplified proxy of the Earth system state that could be enhanced by global state representations. Further investigation will reveal the contexts where this enhancement is most pronounced. 
Future work can exploit time series for both local and global inputs. In fact, it can be tested if global inputs with a temporal component can replace the indices.
Finally, using models that treat the Earth as a system holds immense promise for scientific knowledge discovery, especially when incorporating knowledge of the physical systems.

% The fact that coarsened global views enhance performance suggests that the climatic and oceanic indices are a simplified proxy of the Earth system state that can be enhanced with global state representations. It is interesting to identify the contexts where this enhancement is most pronounced. Methodologically, the approaches can be improved by incorporating knowledge of the physical systems, leading into hybrid modelling. 

% Qualitative Results - Danger Maps

\section{Conclusions}
\label{sec:conclusion}
In this work, we propose TeleViT, a transformer-based architecture that can leverage local and non-local information to model global Earth system processes. Using the SeasFire cube, we showcase our model's ability to improve burned area pattern forecasting using climatic indices and coarsened global views of fire driver variables. This improvement is intensified for longer temporal forecasting horizons, where teleconnections are expected to have a more significant impact, strengthening our hypothesis that TeleViT can discover such long spatio-temporal Earth system interactions.  

The combination of local information, climatic indices, and global views of the Earth holds significant potential for various HADR applications, promising enhanced anticipation capabilities at longer temporal horizons. This includes forecasting extreme events such as floods or droughts, the intensity of tropical cyclones, impacts on food security or human displacement. These applications can benefit from the enhanced prognosis, potentially improving proactive strategies and resource allocation in the face of disasters.

\section*{Acknowledgements}

This work is part of the SeasFire project, which deals with "Earth System Deep Learning for Seasonal Fire Forecasting" and is funded by the European Space Agency (ESA) in the context of the ESA Future EO-1 Science for Society Call.

%% file: 12_appendix.tex
\appendix
% \onecolumn
\label{appendix}

\section{Input Data from the SeasFire cube}
\label{app:datacube}

Table \ref{tab:datacube-variables} shows which variables are used from the SeasFire cube\cite{alonso_seasfire_2022}, along with some pre-processing made for each variable. Total precipitation and population are log-transformed with $log(1+x)$ to follow a less skewed distribution. For more details on the variables, the reader is referred to the cited dataset.

% \usepackage{booktabs}
\begin{table}[htbp]
\centering
\begin{tabular}{@{}lc@{}}
\toprule
\textbf{Full name}                    & \textbf{Pre-processing}               \\ \midrule
\textbf{Local/Global Variables}          &                                \\ \midrule
Mean sea level pressure                  &                                \\
Total precipitation                      & Log-transformed \\
Vapour Pressure Deficit                   &                                \\
Sea Surface Temperature                  &                                \\
Mean Temperature at 2 meters             &                                \\
Surface solar radiation downwards        &                                \\
Volumetric soil water level 1            &                                \\
Land Surface Temperature at day          &                                \\
Normalized Difference Vegetation Index   &                                \\
Population density                       & Log-transformed \\
Cosine of longitude                      & Calculated                     \\
Sine of longitude                        & Calculated                     \\
Cosine of latitude                       & Calculated                     \\
Sine of latitude                         & Calculated                     \\ \midrule
\textbf{Climatic Indices}                &                                \\ \midrule
Western Pacific Index                    &                                \\
Pacific North American Index             &                                \\
North Atlantic Oscillation               &                                \\
Southern Oscillation Index               &                                \\
Global Mean Temperature                  &                                \\
Pacific Decadal Oscillation              &                                \\
Eastern Asia/Western Russia              &                                \\
East Pacific/North Pacific Oscillation   &                                \\
Ni\~no 3.4 Anomaly                         &                                \\
Bivariate ENSO Timeseries                &                                \\ \midrule
\textbf{Target Variable}                 &                                \\  \midrule
Burned Areas from GWIS                   & Made binary                    \\ \bottomrule
\end{tabular}
\caption{Input and target variables used from the SeasFire cube for all settings. The same variables are used for both local and global views.}
\label{tab:datacube-variables}
\end{table}

\section{Model Details and Hyperparameters}
\label{app:hparams}

Models are trained for 50 epochs. We use the cross-entropy loss and the Adam optimizer to train the models. For the U-Net++ model, the initial learning rate is set to 0.001, while for the Transformer models, it is set to 0.0001. The learning rate is reduced on the plateau and the weight decay is set to 0.000001. The model with the lowest validation loss is used for testing. Before entering the models, local and global inputs are normalized. OCIs, which are anomalies are divided by their standard deviation.

The encoders of both ViT and TeleViT consist of $K=8$ layers, with $A=12$ attention heads each, and an embedding dimension $D=768$. For the asymmetric tokenisation, we set $P_l=(1, 16, 16)$, $P_g=(1, 30, 30)$ and $P_i=1$. This means the following:
\begin{itemize}
    \item The local input $x_l$ of size $(14, 80, 80)$ is tokenised spatially in $5\times5$ number of tokens with size $16\times16$. 
    \item The global input $x_g$ of size $(14, 360, 180)$ is tokenised spatially in $12\times6$ number of tokens with size $30\times30$. 
    \item The OCI input $x_i$ of size $(10, 10)$ is tokenised in $10\times10$ number of tokens.
\end{itemize}

\section{Coarsening the SeasFire cube}
\label{app:coarsening}

The SeasFire cube is provided as a xarray-compatible file that is in Zarr format, which makes it easy to coarsen with xarray. All that is needed is to provide an aggregation function for each of the variables and a coarsening factor for each dimension. We use a coarsening factor of 4 along the longitude and latitude dimensions to convert the cube from $0.25^\circ$ to $1^\circ$ spatial resolution and all the input variables are mean-aggregated.

\section{Interactions between the different input datasets}
\label{app:attn_multires}

Figure \ref{fig:attn_multires} shows different interactions between the different input datasets captured by attention.

\newcommand{\mysize}{\fontsize{8}{0.5}\selectfont}
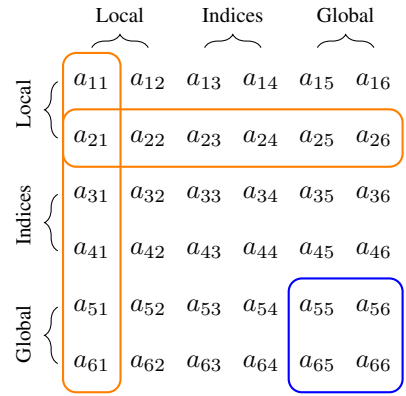
\begin{figure}[htbp]
\centering
\begin{tikzpicture}
 \matrix [nodes={minimum size=7.5mm}]
  {
    \node (a11) {$a_{11}$}; & \node (a12) {$a_{12}$}; & \node (a13) {$a_{13}$};  & \node (a14) {$a_{14}$}; & \node (a15) {$a_{15}$}; & \node (a16) {$a_{16}$};   \\
    \node (a21) {$a_{21}$}; &  \node (a22) {$a_{22}$}; & \node (a23) {$a_{23}$};  & \node (a24) {$a_{24}$}; & \node (a25) {$a_{25}$}; & \node (a26) {$a_{26}$};  \\ 
    \node (a31) {$a_{31}$};& \node (a32) {$a_{32}$}; & \node (a33) {$a_{33}$};  & \node (a34) {$a_{34}$}; & \node (a35) {$a_{35}$}; & \node (a36) {$a_{36}$};   \\
    \node (a41) {$a_{41}$}; & \node (a42) {$a_{42}$}; & \node (a43) {$a_{43}$};  & \node (a44) {$a_{44}$}; & \node (a45) {$a_{45}$}; & \node (a46) {$a_{46}$};   \\
        \node (a51) {$a_{51}$}; & \node (a52) {$a_{52}$}; & \node (a53) {$a_{53}$};  & \node (a54) {$a_{54}$}; & \node (a55) {$a_{55}$}; & \node (a56) {$a_{56}$};   \\
    \node (a61) {$a_{61}$}; & \node (a62) {$a_{62}$}; & \node (a63) {$a_{63}$};  & \node (a64) {$a_{64}$}; & \node (a65) {$a_{65}$}; & \node (a66) {$a_{66}$};   \\
    };
    \draw [decorate, decoration={brace, amplitude=5pt,raise=2pt}, yshift=10pt] (a11.north) -- (a12.north) node [black,above,midway,yshift=8pt] {\mysize Local};
    \draw [decorate, decoration={brace, amplitude=5pt,raise=2pt}, yshift=-2pt] (a13.north) -- (a14.north) node [black,above,midway,yshift=8pt] {\mysize Indices};
    \draw [decorate, decoration={brace, amplitude=5pt,raise=2pt}, yshift=-2pt] (a15.north) -- (a16.north) node [black,above,midway,yshift=8pt] {\mysize Global};

    \draw [decorate, decoration={brace, amplitude=5pt,mirror,raise=2pt}, xshift=-2pt] (a11.west) -- (a21.west) node [black,left,midway,yshift=15pt,xshift=-15pt,rotate=90] {\mysize Local};
    \draw [decorate, decoration={brace, amplitude=5pt,mirror,raise=2pt}, xshift=-2pt] (a31.west) -- (a41.west) node [black,left,midway,yshift=18pt,xshift=-15pt,rotate=90] {\mysize Indices};
    \draw [decorate, decoration={brace, amplitude=5pt,mirror,raise=2pt}, xshift=-2pt] (a51.west) -- (a61.west) node [black,left,midway,yshift=15pt,xshift=-15pt,rotate=90] {\mysize Global};
    \draw[orange, thick,rounded corners=5pt] (a11.north west) rectangle (a61.south east);
    \draw[orange, thick,rounded corners=5pt] (a21.north west) rectangle (a26.south east);
    \draw [blue, thick,rounded corners=5pt] (a55.north west) rectangle (a66.south east);
\end{tikzpicture}
\caption{Depiction of intra-dataset (blue) and inter-dataset interactions (orange) in an attention matrix for a sequence of six tokens with an equal number of tokens for each data source.}
\label{fig:attn_multires}
\end{figure}